%% file: main_body.tex
\documentclass[10pt,twocolumn,letterpaper]{article}

\usepackage{iccv}
\usepackage{times}
\usepackage{epsfig}
\usepackage{graphicx}
\usepackage{amsmath}
\usepackage{amssymb}
\usepackage[dvipsnames]{xcolor}

\usepackage{comment}
\usepackage{mathtools,amsthm}
\usepackage{color, xcolor}
\let\oldproofname=\proofname
\renewcommand{\proofname}{\bf\textit{\oldproofname}}
\usepackage{wrapfig}
\usepackage{multirow}
\usepackage[utf8]{inputenc} 
\usepackage[T1]{fontenc}    
\usepackage{url}            
\usepackage{booktabs}       
\usepackage{amsfonts}       
\usepackage{nicefrac}       
\usepackage{microtype}      
\usepackage{balance}
\usepackage{bm}

\usepackage[pagebackref=true,breaklinks=true,letterpaper=true,colorlinks,bookmarks=false]{hyperref}

\iccvfinalcopy 

\def\iccvPaperID{4015} 

\ificcvfinal\fi

\begin{document}


\title{Towards Novel Target Discovery Through Open-Set Domain Adaptation}

\author{Taotao Jing$^\dag$, Hongfu Liu$^\ddag$, and Zhengming Ding$^\dag$\\
$^\dag$Department of Computer Science, Tulane University, USA\\
$^\ddag$Michtom School of Computer Science, Brandeis University, USA\\
{\tt\small tjing@tulane.edu, hongfuliu@brandeis.edu, zding1@tulane.edu}

}

\maketitle
\ificcvfinal\thispagestyle{empty}\fi

\begin{abstract}
Open-set domain adaptation (OSDA) considers that the target domain contains samples from novel categories unobserved in external source domain. Unfortunately, existing OSDA methods always ignore the demand for the information of unseen categories and simply recognize them as ``unknown'' set without further explanation. This motivates us to understand the unknown categories more specifically by exploring the underlying structures and recovering their interpretable semantic attributes. In this paper,
we propose a novel framework to accurately identify the seen categories in target domain, and effectively recover the semantic attributes for unseen categories. Specifically, structure preserving partial alignment is developed to recognize the seen categories through domain-invariant feature learning. Attribute propagation over visual graph is designed to smoothly transit attributes from seen to unseen categories via visual-semantic mapping. Moreover, two new cross-domain benchmarks are constructed to evaluate the proposed framework in the novel and practical challenge. Experimental results on open-set recognition and semantic recovery demonstrate the superiority of the proposed method over other compared baselines.
\end{abstract}

\section{Introduction}

In recent years, domain adaptation (DA) attracts great interests to address the label insufficiency or unavailability issues, which is the bottleneck to the success of deep learning models \cite{he2016deep}. DA casts a light by transferring existing knowledge from a relevant source domain to the target domain of interest via eliminating the distribution gap across domains \cite{dong2020cscl,panareda2017open}. Most DA efforts focus on the \textit{closed-set domain adaptation} (CSDA) \cite{dong2020cscl,ding2018graph}, assuming the source and target domain share identical label space, which is not always satisfied in real-world scenarios, since the target domain may contain more than we know from source domain. Following this, \textit{open-set domain adaptation} (OSDA) has been widely studied given the source domain only covers a sub-set of the target domain label space\cite{saito2018open, panareda2017open,liu2019separate,kundu2020towards}. Unfortunately, these pioneering OSDA attempts simply identify the known categories while leaving the remaining unobserved samples as an ``unknown'' outlier set. Without any further step, OSDA fails to discover what the unknown categories really are. Interestingly, the target domain may contain some exactly-new categories human beings never see before. This motivates us to further analyze the unknown set more specifically and discover novel categories. 

In this paper, we define such a problem as \emph{Semantic Recovery Open-Set Domain Adaptation} (\textbf{SR-OSDA}), where source domain is annotated with both class labels and semantic attribute annotation, while target domain only contains the unlabeled and unannotated data samples from more categories. The goal of SR-OSDA is to identify the seen categories and also recover the missing semantic information for unseen categories to interpret the new categories in target domain. To our best knowledge, this is a completely new problem in literature with no exploration. The challenges now become two folds: (1) how to accurately identify seen and unseen categories in target domain with well-labeled source knowledge; (2) how to effectively recover the missing attributes of unseen categories.

To this end, we propose a novel framework to simultaneously recognize the known categories and discover new categories from target domain as well interpret them at the semantic level. The general idea of our model is to learn domain-invariant visual features by mitigating the cross-domain shift, and consequently build visual-semantic projection to recover the missing attributes of unknown target categories. Our contributions are highlighted as follows:\vspace{-2mm}
\begin{itemize}
    \item We are the first to address the SR-OSDA problem, and propose a novel and effective solution to identify seen categories and discover unseen one.
    \vspace{-2mm}
    \item We propose structure preserving partial alignment to mitigate the domain shift when target covers larger label space than source, and attributes propagation over visual graph to seek the visual-semantic mapping for better missing attribute recovery.
    \vspace{-2mm}
    \item Two new benchmarks are built for SR-OSDA evaluation.  Our proposed method achieves promising performance in both target sample recognition and semantic attribute recovery.
\end{itemize}

\section{Related Work}
Here we introduce the related work along the open-set domain adaptation and zero-shot learning, and highlight the differences between our work and the existing literature.

\noindent\textbf{Open-Set Domain Adaptation.} Compared to classic closed set domain adaptation \cite{yang2020bi, xia2020structure,xia2020hgnet, chen2020adversarial,tang2020discriminative,zhang2020label,tang2020unsupervised, jing2020adaptively, jing2021adversarial, wang2020ev}, open-set domain adaptation manages a more realistic task when the target domain contains data from classes never present in the source domain~\cite{bucci2020effectiveness,rakshit2020multi,luo2020progressive,pan2020exploring, kundu2020towards, feng2019attract, tan2019weakly,baktashmotlagh2019learning, saito2018open}. Busto \etal attempts to study the realistic scenario when the source and target domain both includes exclusive classes from each other  \cite{panareda2017open}. Later on, Saito \etal focus on the situation when the source domain only covers a subset of the target domain label space and utilizes adversarial framework to generate features and recognizes samples deviated from the pre-defined threshold as ``unknown'' \cite{saito2018open}. Instead of relying on the manually pre-defined threshold, \cite{feng2019attract} takes advantage of the semantic categorical alignment and contrastive mapping to encourage the target data from known classes to move close to corresponding centroid while staying away from unknown classes. STA adopts a coarse-to-fine weighting mechanism to progressively separate the target data into known and unknown classes \cite{liu2019separate}. Most recently,  SE-CC augments the Self-Ensembling technique to with category-agnostic clustering in the target domain \cite{pan2020exploring}.

\noindent\textbf{Zero-shot learning.} Demand of leveraging annotated data to recognize novel classes unseen before motivates a boom thread of research known as \textit{Zero-Shot Learning} (ZSL) \cite{ding2017low, vyas2020leveraging,chen2020boundary,xie2020region,wan2019transductive,yu2019zero,elhoseiny2019creativity,ding2019marginalized,jiang2019transferable}. Early ZSL works explore class semantic attributes as intermediate to classify the data from unseen classes \cite{lampert2013attribute,lampert2009learning}. Some ZSL methods learns a mapping between the visual and semantic spaces to compensate for the lack of visual features from the unseen categories \cite{chao2016empirical,akata2015label}. However, ZSL methods do not guarantee the discrimination between the seen and unseen classes, leading to bias towards seen classes under another realistic scenario, \textit{Generalized Zero-Shot Learning} (GZSL). GZSL assumes the target data to evaluate are drawn from the whole label space including seen and unseen classes \cite{jiang2019transferable,schonfeld2019generalized,Huang_2019_CVPR,chao2016empirical,felix2018multi}. Recently, generative frameworks are explored to generate synthesized visual features from unseen classes boosts the performance of ZSL and GZSL \cite{wen2016discriminative, wang2018zero}. \cite{zhu2018generative,wen2016discriminative} use a Wasserstein GAN~\cite{arjovsky2017wasserstein} and the seen categories classifier to increase the discrimination of the synthesized features. \cite{felix2018multi} utilizes the cycle consistency loss to optimize the synthesized feature generator, and 
\cite{wang2018zero} 
study conditional VAEs \cite{kingma2013auto} to learn the feature generator. 

Different from open-set domain adaptation, the proposed SR-OSDA problem demands to recover interpretable knowledge of the target data from classes never present in the source domain, and uncover new classes. Moreover, SR-OSDA is more challenging than the GZSL problem because we do not have access to the attributes nor any other semantic knowledge of the target domain new categories, which makes SR-OSDA a more realistic and practical problem.

\begin{figure*}
\centering
  \includegraphics[width=0.98\textwidth]{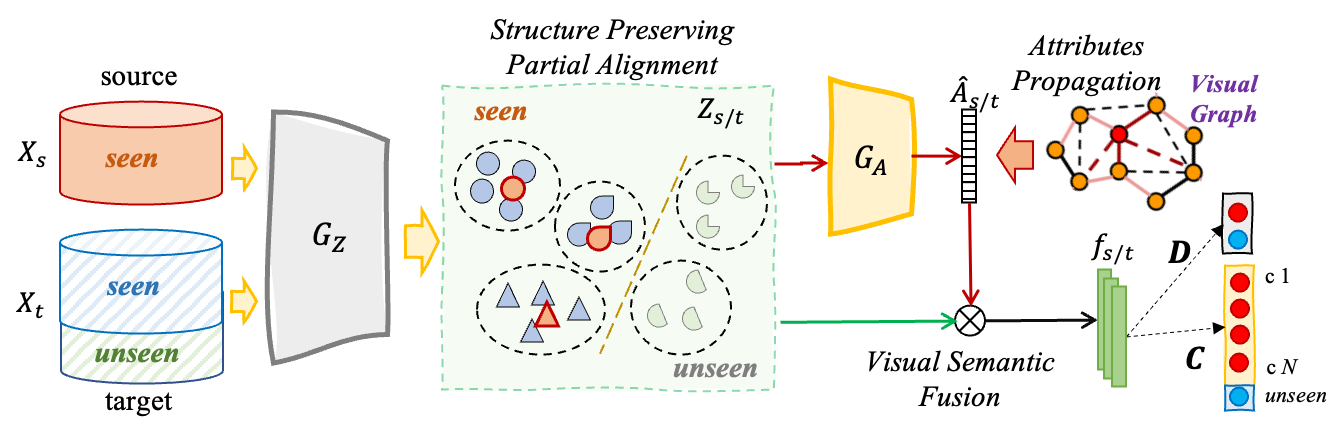}\vspace{-3mm}
  \caption{Illustration of our proposed framework, where $X_t$ contains some unseen categories from $X_s$. Convolutional neural networks (e.g., ResNet \cite{he2016deep}) are used as backbone to extract visual features $X_{s/t}$, which are further input to $G_Z$ to learn domain-invariant features $Z_{s/t}$ through partial alignment. $G_A$ then maps $Z_{s/t}$ to semantic attributes $A_s$. Visual-semantic features are fused for the final classification tasks, one is $D$ to identify seen/unseen from target data, and the other $C$ to recognize all cross-domain data into $K_s$+1 classes (i.e., $K_s$ seen + one unseen large category). } \label{fig:framework}\vspace{-6mm}
\end{figure*}

\input{notation}

\section{Motivations and Problem Definition}
In this section, we illustrate our motivations and provide the problem definition of the semantic recovery open-set domain adaptation.

Open-set domain adaptation tasks \cite{panareda2017open} focus on the scenario when the target domain contains data from classes never observed in the source domain, which is more practical than the conventional closed-set domain adaptation \cite{dong2020cscl}. However, existing open-set domain adaptation efforts simply identify those unseen target samples as one large unknown category and give up exploring the discriminative and semantic knowledge inside the unknown set. The demand of further understanding the novel classes that only exist in the target domain motivates us to study how to recover missing semantic attributes to explain the target data and discover novel classes, which leads to the problem \emph{Semantic Recovery Open-Set Domain Adaptation} (\textbf{SR-OSDA}) addressed in this paper. The main challenges of SR-OSDA lie in not only identifying the target samples in the unseen classes, but also providing the partitional structures of these samples with recovered semantic attributes for further interpretation. 

For better understanding, we clarify the problem with mathematical notations. The target domain is defined as $\mathcal{D}_t = \{\mathbf{X}_{t}\}$ containing $N_t$ samples with visual features from $K_t$ categories. The auxiliary source domain $\mathcal{D}_s = \{\mathbf{X}_{s}, \mathbf{Y}_{s}, \mathbf{A_s}\}$ consists of $N_s$ samples from $K_s$ classes with visual features $\mathbf{X}_{s}$, labels $\mathbf{Y}_s$, and semantic attributes $\mathbf{A}_s$. For each source sample, the semantic attributes $\mathbf{a}_s^i = \mathcal{A}^{y_s^i}, \mathbf{a}_s^i \in \mathbb{R}^{d_\mathbf{a}}$ are obtained from $\mathcal{A}$, which consists of class-wise attributes of the source domain. SR-OSDA aims to recover the missing semantic attributes for the target data based on the visual features, and uncover novel categories never present in the source domain. Table~\ref{table:notation} shows several key notations and descriptions in the SR-OSDA setting.

It is noteworthy that the source and target domains are drawn from different distributions. Besides, the target data set covers all classes in the source domain, as well as $K$ exclusive categories only exist in the target domain, where $K = K_t - K_s > 0$.
SR-OSDA is different from open-set domain adaptation, which ignores to recover interpretable knowledge and discover new classes in the target domain. Moreover, the defined problem is different from generalized zero-shot learning~\cite{schonfeld2019generalized}, as we have no access to the semantic knowledge of the target domain unseen categories.

To our best knowledge, SR-OSDA is the first time proposed, aiming to discover novel target classes via recovering semantic attributes from the auxiliary source data. In the following, we illustrate our solution to learn the relationship between the visual features and semantic attributes with the guidance of the source data, which can be transferred to the target data and interpretably discover unseen classes.

\section{The Proposed Method}

\subsection{Framework Overview}

To tackle the above SR-OSDA problem, we propose a novel target discovery framework (Figure \ref{fig:framework}) to simultaneously recognize the target domain data from categories already observed in the source domain, and recover the interpretable semantic attributes for the unknown target classes from the source. To achieve this, three modules are consequently designed to address the cross-domain shift, semantic attributes prediction and task-driven open-set classification. Specifically, the source data are adapted to the target domain feature space through partial alignment while preserving the target structure. A projector $G_A(\cdot)$ bridging the domain invariant feature space $\mathbf{z}_{s/t}^i$ and the semantic attributes space $\mathbf{a}_{s/t}^i$ is trained by the source data as well as the target data with confident pseudo attributes. Moreover, the visual features will guide the attributes propagated from seen categories to unseen ones, and the semantic attributes will also promote the visual features discrimination through joint visual-semantic representation recognition for $C(\cdot)$ and $D(\cdot)$, where $D(\cdot)$ is a binary classifier to identify seen and unseen target samples, and $C(\cdot)$ is an extended multi-class classifier with $K_s+1$ outputs. 

Since the target data are totally unlabeled and all three modules rely on the label information in target domain, we first discuss how to obtain the pseudo labels of target samples through our design progressive seen-unseen separation stage. That is, we will assign target samples into $K_s$ observed categories and $K$ unobserved categories. 
In the following, we introduce the progressive seen-unseen separation and three key modules in our proposed framework.

\subsection{Modules and Objective Function}

\noindent\textbf{Progressive Seen-Unseen Separation}. Here we describe the initialization strategy to separate the target domain data into seen and unseen sets based on the visual features space. Intuitively, part of source-style target samples are promisingly identified by the well-trained source model, which are actually belonging to seen categories more probably. On the other hand, those target samples assigned with even and mixed prediction probabilities across multiple classes tend to be unseen categories, as no source classifier can easily recognize them. To achieve this, we apply the prototypical classifier to measure the similarities between each target sample to all source class prototypes \cite{snell2017prototypical}. For each target sample $\mathbf{x}_t^i$ and the source $K_s$ prototypes $\{\mu^c|_{c=1}^{K_s}\}$, the probability prediction is defined as:
\begin{equation}
    \label{eq:prot}
    p(y_t^i = c | \mathbf{x}_t^i) = \frac{\exp{\big(-d(\mathbf{x}_t^i, \,\, \mu^c)\big)}}{\sum_{c'}\exp\big({-d(\mathbf{x}_t^i, \,\, \mu^c)}\big)},
\end{equation}
where $d(\cdot)$ is the distance function. The highest probability prediction $p_t^{i}$ is adopted as the pseudo label $\tilde{y}_t^{i}$ for $\mathbf{x}_t^i$. Next, we adopt a threshold $\tau$ to progressively separate all target samples into seen $\mathcal{D}_t^{s}$ and unseen sets $\mathcal{D}_t^{u}$. The number of samples in $\mathcal{D}_t^{s}$ and $\mathcal{D}_t^{u}$ are denoted as $N_t^s$ and $N_t^u$, respectively. 
Specifically, we define $\tau$ the mean of the highest probability prediction of all target samples, i.e., $\tau = \frac{1}{N_t} \sum_{\mathbf{x}_t^i \in \mathcal{D}_t} p_t^i$. Based on that, we can build two sets as:
\begin{equation}
    \label{eq:separate}
    \left\{\begin{matrix}
     \mathbf{x}_t^i \in \mathcal{D}_t^{s}, & p_t^i \geq \tau\\ 
     \mathbf{x}_t^i \in \mathcal{D}_t^{u}, & p_t^i < \tau
    \end{matrix}\right. .
\end{equation}

Since we only have the source prototypes in the beginning, they are not accurate to identify seen and unseen sets due to the domain shift. Thus, we can gradually update the seen prototypes by involving newly-labeled target samples from $\mathcal{D}_t^{s}$ as $\mu^c = (1-\alpha)\mu^c + \alpha \frac{1}{N_t^{s(c)}}\sum_{\mathbf{x}_t^i \in \mathcal{D}_t^{s(c)}}\mathbf{x}_t^i$, where $\mathcal{D}_t^{s(c)}$ denotes a set of $N_t^{s(c)}$ target samples predicted as $\tilde{y}_t^i=c$ confidently, and $\alpha$ is the small value to control the mixture of cross-domain prototypes. 

After obtaining all pseudo labels in the seen set $\mathcal{D}_t^{s}$, we also need to explore more specific knowledge in $\mathcal{D}_t^{u}$ instead of treating it as a whole like OSDA \cite{saito2018open}. 
To this end, we apply K-means clustering algorithm to group $\mathcal{D}_t^{u}$ into $K$ clusters with the cluster center as $\{\eta^{k_1}, \cdots, \eta^{K}\}$. In this way, we can obtain all prototypes of \textit{seen} and \textit{unseen} categories as ${\mathcal{R}_\mathbf{x}} = \{ \mu^1, \cdots, \mu^{K_s}, \eta^{k_1}, \cdots, \eta^{K}\}$. In order to refine the pseudo labels of target samples, we adopt K-means clustering algorithm with centers initialized as ${\mathcal{R}_\mathbf{x}}$ over $X_t$ until the results are converged. 

To this end, we obtain all pseudo labels for target samples. We also assign semantic attribute to seen target samples based on their pseudo label belonging to which source category. Next, we explore structure preserving partial alignment, attribute propagation and task-driven classification to solve SR-OSDA.

\vspace{1mm}\noindent\textbf{Structure Preserving Partial Alignment}. Due to the disparity between the source and target domains label spaces, directly matching the feature distribution across domains is destructive. Considering our goal of uncovering the \textit{unseen} categories in the target domain, preserving the structural knowledge of the target domain data becomes even more crucial. Thus, instead of mapping the source and target domains into a new domain-invariant feature space, we seek to align the source data to the target domain distribution through partial alignment. 


Specifically, with the help of the target domain pseudo labels $\mathbf{\tilde{Y}}_t$, for each class $c$ in the pseudo label space, which contains $K_s+K$ categories, the prototype can be calculated as the class center in the space of feature $\mathbf{z}$ can be calculated as $\mathcal{R}_{\mathbf{z}}^c = \mathbb{E}_{\mathbf{x}_t^i\in\mathcal{D}_t} \mathbf{z}_t^i \cdot \mathbf{1}_{\tilde{y}_t^i=c}$. The prototypes $\mathcal{R}_{\mathbf{z}}$ describe the class-wise structural knowledge in the target domain in the $\mathbf{z}$ feature space. To solve the domain disparity, we align each source sample to its specific target center and also keep away from other target centers as:
\begin{equation}
    \label{eq:loss_s_cent}
    \begin{aligned}
    \mathcal{L}_{s}^{R} = & \frac{1}{N_s} \sum_{i=1}^{N_s}\sum_{c=1}^{|\mathcal{R_{\mathbf{z}}}|} \Big(\mathbf{1}_{y_s^i=c}d(\mathbf{z}_s^i, \mathcal{R}_{\mathbf{z}}^c) 
    - \frac{\mathbf{1}_{y_s^i \neq c}}{|\mathcal{R_{\mathbf{z}}}|-1} d(\mathbf{z}_s^i, \mathcal{R}_{\mathbf{z}}^{c})\Big),
    \end{aligned}
\end{equation}
where $K_s + K = |\mathcal{R_{\mathbf{z}}}|$ is the total number of prototypes in $\mathcal{R}_{\mathbf{z}}$. Moreover, we deploy the similar loss to make within-class target samples more compact while keeping between-class target samples more discriminative as:
\begin{equation}
    \label{eq:loss_t_cent}
    \begin{aligned}
    \mathcal{L}_{t}^{R} = & \frac{1}{N_t} \sum_{i=1}^{N_t} \sum_{c=1}^{|\mathcal{R_{\mathbf{z}}}|} \Big(\mathbf{1}_{\tilde{y}_t^i=c} d(\mathbf{z}_t^i, \mathcal{R}_{\mathbf{z}}^c)- \frac{\mathbf{1}_{\tilde{y}_t^i \neq c}}{|\mathcal{R_{\mathbf{z}}}|-1} d(\mathbf{z}_t^i, \mathcal{R}_{\mathbf{z}}^{c})).
    \end{aligned}
\end{equation}
Such a loss function will make within-class target samples more compact while pushing away from others.


These two loss functions help align source and target to obtain domain-invariant visual features and also seek more discriminative knowledge over target samples. Then we obtain the objective of structure preserving partial domain adaptation as $\mathcal{L}^{R} = \mathcal{L}_{s}^{R} + \mathcal{L}_{t}^{R}$.

\vspace{1mm}\noindent\textbf{Attributes Propagation with Visual Structure}. Since unseen target samples are totally without any annotations either class label or semantic attributes, our goal is to recover their semantic attributes via visual-semantic projector $G_A(\cdot)$. However, only attributes knowledge of the classes seen in the source domain is available for training, while the target samples from unseen categories have no way to optimize the $G_A(\cdot)$, which might lead the projector $G_A(\cdot)$ towards bias to the seen categories when dealing with unseen target class samples. To this end, we propose the mechanism of attributes propagation to aggregate the visual graph knowledge into the semantic description projection, which is beneficial to the attributes propagated from seen classes to unseen classes.

Specifically, for features $\mathbf{z}^i = G_Z(\mathbf{x}^i)$ of a training batch, the adjacency matrix $A$ is calculated as $A_{ij}=\exp(-{d_{ij}^2}/{\sigma^2})$, where $A_{ii}=0, \forall i$, and $d_{ij} =\| \mathbf{z}^i - \mathbf{z}^j\|_2$ is the distance of $(\mathbf{z}^i, \mathbf{z}^j)$. $\sigma$ is a scaling factor set as $\sigma^2 = \mathrm{Var}(d_{ij}^2)$ as \cite{rodriguez2020embedding} to stabilize training. The attributes projected from visual features are reconstructed as:
\begin{equation}
    \label{eq:att_prop}
    \mathbf{\hat{a}}^i = \sum\nolimits_{j} {W}_{ij} G_A\Big(G_Z(\mathbf{x}^j)\Big),
\end{equation}
where $L = D^{-\frac{1}{2}} A D^{-\frac{1}{2}}, \,\, D_{ii} = \sum_j A_{ij}$ and ${W}=(\mathbf{I} - \beta L)^{-1}$, in which $\beta \in \mathbb{R}$ is a scaling factor fixed as suggested by \cite{rodriguez2020embedding}, and $\mathbf{I}$ is the identity matrix. After the semantic attributes propagation, $\mathbf{\hat{a}}_{s/t}^i$ is refined as a weighted combination of its neighbors guided by the visual graph. This benefits attributes projector from overfitting to the \textit{seen} categories, while removing undesired noise \cite{rodriguez2020embedding}. 

After the projected attributes refinement via attribution propagation, we optimize the attributes projector $G_A(\cdot)$ on the seen categories across two domains as:
\begin{equation}
    \label{eq:loss_s_att}
    \mathcal{L}^A = \frac{1}{N_{s}+N_{t}^s} \sum_{\mathbf{x}^i \in \mathcal{D}_s \cup \mathcal{D}_t^s} L_{bce} (\hat{\mathbf{a}}^i, \,\,\, \mathbf{a}^i),
\end{equation} 
where $L_{bce}(\cdot)$ is the binary cross-entropy loss, and $N_{t}^s$ is the number of samples in $\mathcal{D}_t^s$. Each dimension of the semantic attributes $\mathbf{a}^i \in \mathbb{R}^{d_\mathbf{a}}$ represents one specific semantic characteristic, and $\hat{\mathbf{a}}^i$ describes the predicted probability that the input sample has specific characteristics. 




\vspace{1mm}\noindent\textbf{Visual-Semantic Fused Recognition}. Since visual features and semantic attributes describe the data distribution from different perspectives. To simultaneously leverage the multi-modality benefits of visual and semantic descriptions, we explore the joint visual and semantic representation by conveying the semantic discriminative information $\mathbf{a}^i$ into the visual feature $\mathbf{z}^i$ as $\mathbf{f}^i =  \mathbf{z}^i \oplus \mathbf{a}^i$,
where $\oplus$ is concatenating $\mathbf{z}^i$ and $\mathbf{a}^i$ as joint feature $\mathbf{f}^i$. 

It is noteworthy that during the training, several different semantic attributes are available in different stages, e.g., ground-truth ($\mathbf{a}^i$), pseudo attributes ($\mathbf{\tilde{a}}^i$), and predicted attributes ($\mathbf{\hat{a}}^i$). We take them all into account and will obtain various joint representations as:
\begin{equation}
\label{eq:f_list}
\left\{\begin{matrix*}[l]
 \mathcal{F}_s^i = \{\mathbf{f}_s^i, \mathbf{\hat{f}}_{s}^i\} ,& \mathbf{x}_s^i \in \mathcal{D}_s\\ 
 \mathcal{F}_t^{i} = \{\mathbf{\tilde{f}}_t^i, \mathbf{\hat{f}}_t^i \} ,&  \mathbf{x}_t^i \in \mathcal{D}_t^s \\
 \mathcal{F}_t^{i}=\{ \mathbf{\hat{f}}_{t}^i\} ,& \mathbf{x}_{t}^i \in \mathcal{D}_{t}^u
\end{matrix*}\right. ,
\end{equation}
where $\mathbf{f}_s^i = \mathbf{z}_s^i \oplus \mathbf{a}_s^i$,  $\mathbf{\tilde{f}}_t^i = \mathbf{z}_t^i \oplus \mathbf{\tilde{a}}_t^i$, and $\mathbf{\hat{f}}_{s/t}^{i} = \mathbf{z}_{s/t}^{i} \oplus \mathbf{\hat{a}}_{s/t}^{i}$. All joint features in $\mathcal{F}_s$ and $\mathcal{F}_t$ are input into the classifier $C(\cdot)$ and $D(\cdot)$ to optimize the framework.

To maintain the performance of classifier $C(\cdot)$ over supervision from source and target domains, we construct the cross-entropy classification loss as:
\begin{equation}
    \label{eq:loss_s_ce}
    \mathcal{L}^C = \frac{1}{N_{s}+N_{t}} \sum_{\mathbf{f}^i \in \mathcal{D}_s \cup \mathcal{D}_t } L_{ce} (C(\mathbf{f}^i), y^i),
\end{equation}
where $L_{ce}(\cdot)$ is the cross-entropy loss and $y^i$ denotes the $K_s$ source labels and $K_s+1$ target labels. Moreover, we train a binary classifier $D(\cdot)$ to separate the target domain into \textit{seen} and \textit{unseen} subsets, which can be optimized by:
\begin{equation}
    \label{eq:loss_st_bi}
    \begin{aligned}
    \mathcal{L}_{t}^{D} = \frac{1}{N_t} \sum_{\mathbf{x}_t^i \in \mathcal{D}_t} \sum_{\mathbf{f} \in \mathcal{F}_t^{i}} L_{bce} (D(\mathbf{f}), \,\, \psi(\tilde{y}_t^i)),
    \end{aligned}
\end{equation}
in which $\psi(\tilde{y}_t^i)$ indicates if the target sample $\mathbf{x}_t^i$ is from the \textit{seen} categories ($\psi(\tilde{y}_t^i)=0, \,\, \mathbf{x}_t^i \in \mathcal{D}_t^s$), or from the \textit{unseen} categories ($\psi(\tilde{y}_t^i)=1, \,\, \mathbf{x}_t^i \in \mathcal{D}_t^u$).



Then we have our classification supervision objective on both source and target domain with joint visual and semantic representations as $\mathcal{L}^{T} = \mathcal{L}^{C} + \mathcal{L}_{t}^{D}$.

\vspace{1mm}\noindent\textbf{Overall Objective Function}. To sum up, we can obtain the overall objective function by integrating the structure preserving partial adaptation, semantic attributes propagation and prediction, and joint visual-semantic representation recognition as:
\begin{equation}
    \label{eq:obj_overall}
    {\underset{G_Z, G_A, C, D}{\min}} \mathcal{L}^{T}+\lambda_1 \mathcal{L}^R + \lambda_2 \mathcal{L}^A,
\end{equation}
where $\lambda_1$ and $\lambda_2$ are two trade-off parameters. Through minimizing the proposed objective, the semantic descriptive knowledge is aggregated from the source data into the unlabeled target domain through the joint visual-semantic representation supervision and attributes propagation. Meanwhile, the discriminative visual structure in the target domain is promoted by the cross-domain partial adaptation.

\input{dataset}

\input{D2AwA_open}

\input{D2AwA_semantic}

\section{Experiments}

\subsection{Experimental Settings}

\noindent \textbf{Datasets}. We construct two datasets for the novel SR-OSDA setting. (1) \textit{D2AwA} is constructed from the DomainNet dataset \cite{peng2019moment} and AwA2\cite{xian2017zero}. Specifically, we choose the shared 17 classes between the DomainNet and AwA2, and select the alphabetically first 10 classes as the seen categories, leaving the rest 7 classes as unseen. The corresponding attributes features in AwA2 are used as the semantic description. It is noteworthy that DomainNet contains 6 different domains, while some of them barely share the semantic characteristics described by the attributes of AwA2, e.g., quick draw. Thus, we only take the ``real image'' (R) and ``painting'' (P) domains into account, together with the AwA2 (A) data for model evaluation. (2) \textit{I2AwA} is collected by \cite{zhuo2019unsupervised} consisting of 50 animal classes, and split into 40 seen categories and 10 unseen categories as \cite{xian2017zero}. The source domain (I), includes 2,970 images from seen categories collected via Google image search engine, while the target domain comes from AwA2 (Aw) dataset for zero-shot learning with 37,322 images in all 50 classes \cite{xian2017zero}. We use the binary attributes of AwA2 as the semantic description, and only the seen categories attributes of source data are available for training. Only one task I$\rightarrow$Aw is evaluated on \textit{I2AwA}. Table~\ref{table:data} shows several statistical characteristics of \textit{D2AwA} and \textit{I2AwA}.

\vspace{1mm}\noindent\textbf{Evaluation Metrics}. We evaluate our method in two aspects: (1) target sample recognition under the open-set domain adaptation and (2) generalized semantic attribute recovery. For the first one, we follow the conventional open-set domain adaptation studies~\cite{panareda2017open,saito2018open}, recognizing the whole target domain data into one of the seen categories or ``unknown'' category. The standard open-set domain adaptation average accuracy calculated on all the classes are reported as OS. Besides, we report the average accuracy calculated on the target domain seen classes as OS$^*$, while for the target domain unseen categories, the accuracy is reported as OS$^\diamond$.
For semantic attribute recovery, we compare the predicted semantic description with the ground-truth semantic attributes. Specifically, we adopt a TWO-stage test: (a) identifying a test sample from \textit{seen} or \textit{unseen} set, (b) applying prototypical classification with corresponding \textit{seen}/\textit{unseen} ground-truth attributes. We report the performances on the seen categories and unseen categories as $S$ and $U$, respectively, and calculate the harmonic mean $H$~\cite{schonfeld2019generalized}, defined as $H=2\times S\times U/(S + U)$. Note that all results we reported are the average of class-wise top-1 accuracy, to eliminate the influence caused by the imbalanced class. 

\noindent\textbf{Implementation}. We use the pre-trained ResNet-50~\cite{he2016deep} on ImageNet as the backbone, and take the second last fully connected layer as the features $\mathbf{X}_{s/t}$ \cite{deng2009imagenet, he2016deep}. $G_Z(\cdot)$ is a two-layer fully connected neural network with hidden layer dimension as 1,024, and the output feature dimension is 512. $C(\cdot)$ and $D(\cdot)$ are both two-layer fully connected neural networks classifier with hidden layer dimension as 256, and the output dimension of $C(\cdot)$ is $K_s+1$, while the output of $D(\cdot)$ is just two dimensions indicating seen or unseen classes. $G_A(\cdot)$ is a two-layer neural network with hidden layer dimension as 256 followed, and the final output dimension is the same as the semantic attributes dimension followed by Sigmoid function. We employ the cosine distance for the prototypical classification, while all other distances used in the paper are Euclidean distances. For simplicity, we adopt the ground-truth novel classes number as $K$, and we notice that the results are not sensitive to the value of $K$ within a range. There are many cluster number estimation methods but out of scope in this work. For parameters, we fix $\alpha=0.001$, $\beta=0.2$, $\lambda_1=10^{-4}$, $\lambda_2 = 0.1$, and the learning rate is fixed as $10^{-3}$ for all experiments, and report the 100-th epoch results for all the experiments. Source code of this work is available online\footnote{\url{https://github.com/scottjingtt/SROSDA.git}}.

\begin{figure*}\vspace{-4mm}
\centering
  \includegraphics[width=1\textwidth]{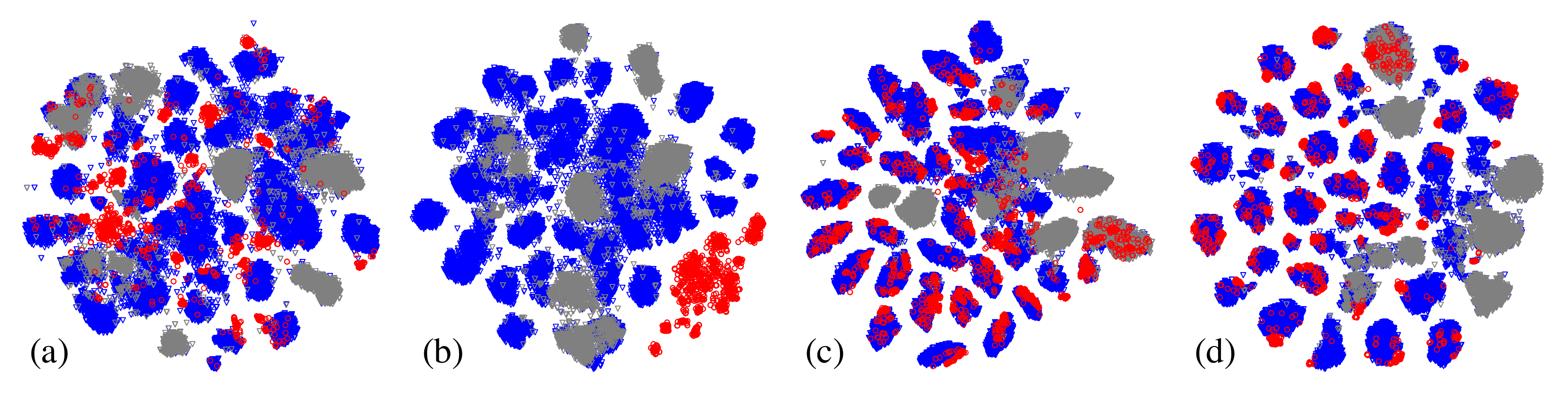}\vspace{-2mm}
  \caption{tSNE visualization of representations generated by (a) ResNet, (b) STA, and (c) Ours on \textit{I2AwA}. (d) shows the joint visual-semantic features proposed in our paper. \textcolor{red}{Red} circles denote source data. \textcolor{blue}{Blue} and \textcolor{gray}{gray} triangles denote target domain seen and unseen classes. }\label{fig:tsne}\vspace{-4mm}
\end{figure*}

\begin{figure}
\centering
  \includegraphics[width=0.48\textwidth]{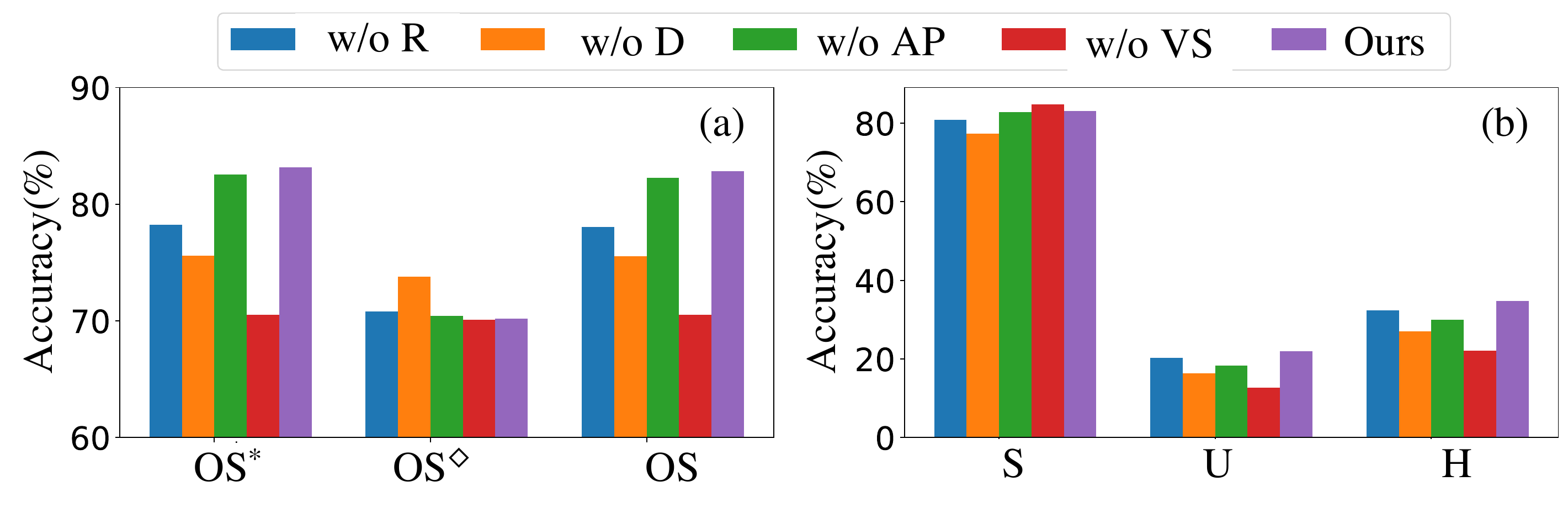}\vspace{-2mm}
  \caption{Ablation study of our proposed model on \textit{I2AwA} by removing specific one of structure preserving partial alignment (w/o $\mathcal{L}^R$), binary classifier (w/o $\mathcal{L}^D$,) attributes propagation (w/o AP), or joint visual-semantic representation (w/o VS).}\label{fig:terms}\vspace{-4mm}
\end{figure}

\begin{figure*}
\centering
  \includegraphics[width=1\textwidth]{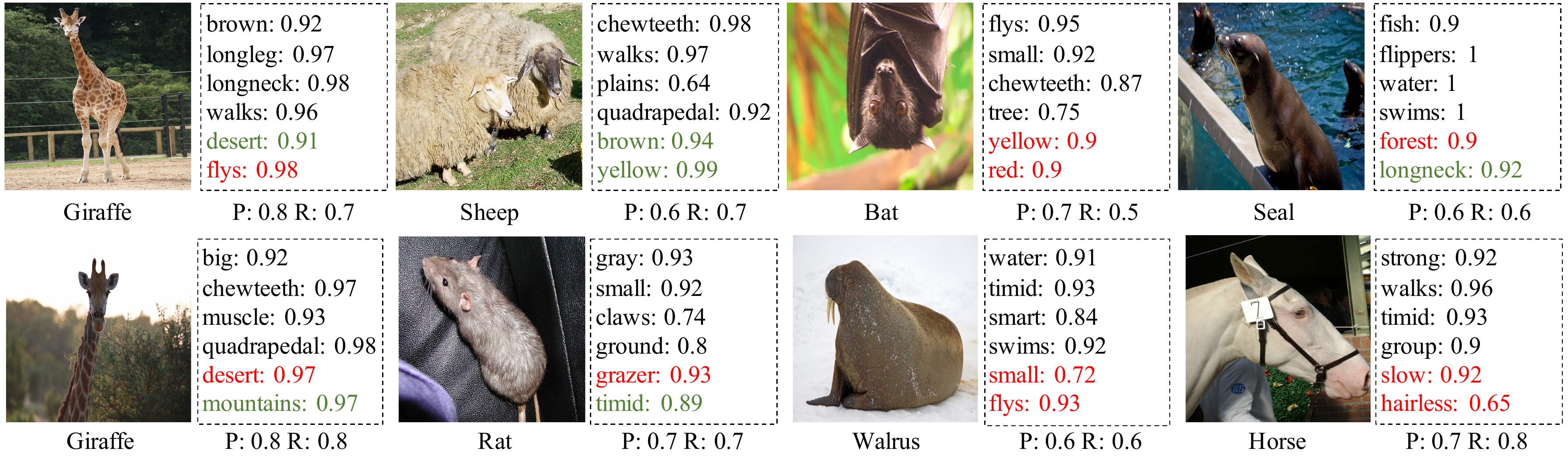}
  \caption{Selected samples from AwA2 dataset and attributes predicted by our method. The \textbf{black} ones are correctly predicted attributes, \textcolor{red}{red} ones are wrong prediction, and the \textcolor{OliveGreen}{green} ones are wrong predictions but reasonable for the specific instance. ``P'' and ``R'' denote precision and recall of the attributes prediction for each sample, respectively.} \label{fig:samples} \vspace{-6mm}
\end{figure*}

\begin{figure}
\centering
  \includegraphics[width=0.5\textwidth]{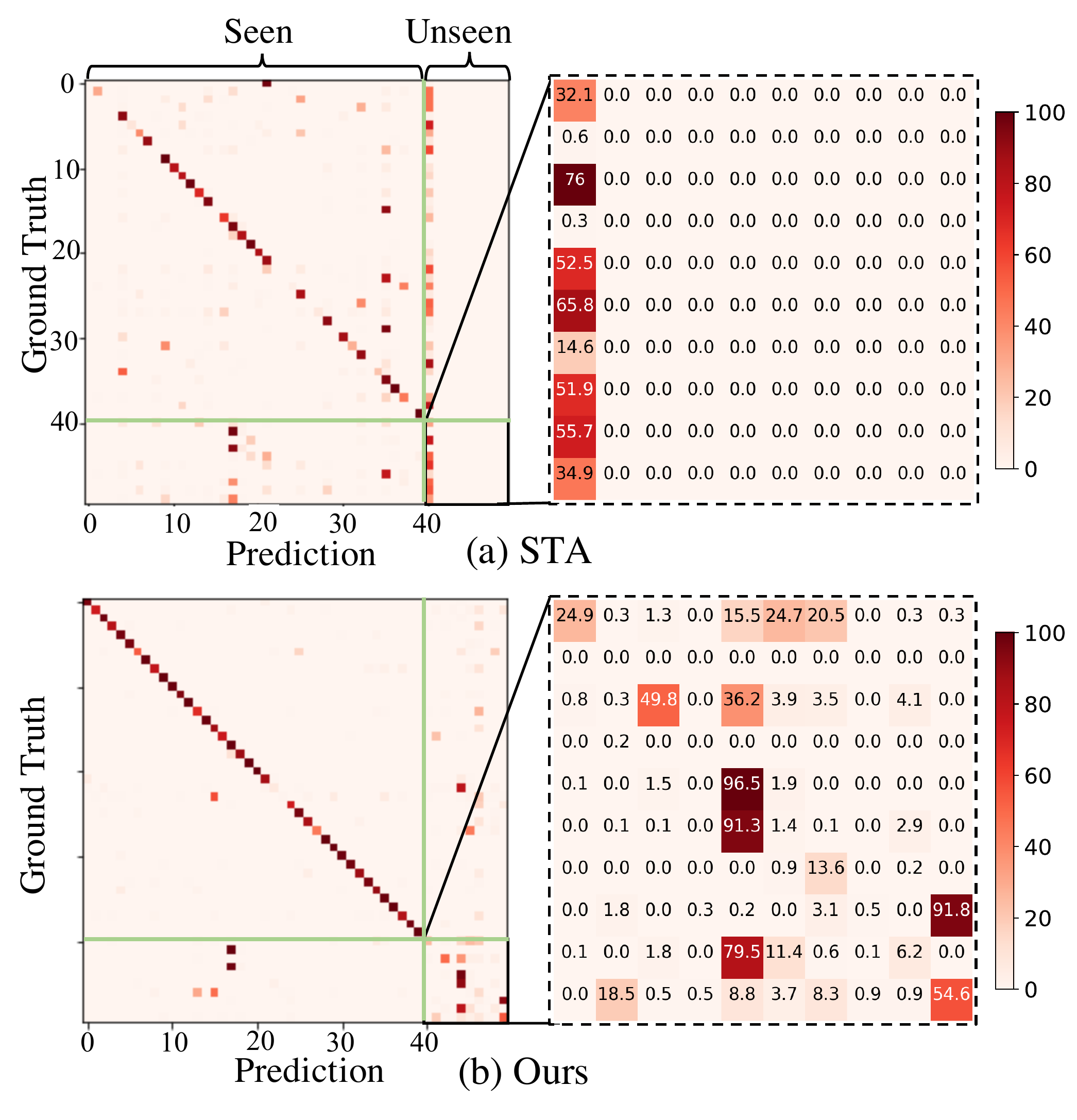}\vspace{-2mm}
  \caption{Confusion matrix of target samples from \textit{I2AwA}. (a) shows the results of STA and (b) lists ours. The unseen classes are zoomed in for better visualization.} \label{fig:confusion} \vspace{-6mm}
\end{figure}

\vspace{1mm}\noindent \textbf{Competitive Methods}. Since the problem we address in this paper is in a novel and practical setting, we mainly compare two distinctive branches of baselines in terms of open-set domain adaptation and zero-shot learning.

For open-set domain adaptation, we compare our method with OSBP \cite{saito2018open}, AOD \cite{feng2019attract}, and STA \cite{liu2019separate}. OSBP utilizes the adversarial training strategy to extract features for the target data, which is recognized into seen/unseen classes by a pre-defined threshold \cite{saito2018open}. AOD exploits the semantic structure of open set data from categorical alignment and contrastive mapping to push the unknown classes away from the decision boundary \cite{feng2019attract}. Differently, STA adopts a coarse-to-fine mechanism to progressively separate the known and unknown data without any manually set threshold \cite{liu2019separate}. 

For the semantic recovery tasks, we implement a source-only trained neural network, and two zero-shot learning methods, ABP \cite{zhu2019learning} and TF-VAE \cite{narayan2020latent} under our setting, as baselines. The source-only model is a fully-connected neural network trained with only source domain ResNet-50 \cite{he2016deep} features available, which learns a projector mapping the visual features to semantic attributes. ABP trains a conditional generator mapping the class-level semantic features and Gaussian noise to visual features \cite{zhu2019learning}. TF-VAE propose to enforce semantic consistency at all training, feature synthesis, and classification stages \cite{narayan2020latent}. Besides, both ABP and TF-VAE are able to handle generalized zero-shot learning problems given the semantic attributes from the whole target label space. We also report ABP* and TF-VAE*, which take extra the semantics of unseen target categories as inputs.

\subsection{Algorithmic Performance}
Table~\ref{table:D2AwA_open} shows the open-set domain adaptation accuracy on \textit{D2AwA} and \textit{I2AwA}. From the results we observe that our proposed method outperforms all compared baselines in terms of overall accuracy on most tasks. Especially on the task A$\rightarrow$R, our model improves $10.4\%$ over the second best compared method. The significant improvements come from our effective framework and the extra source semantic information. Note that in the classical open-set domain adaptation, none of the semantic attributes are leveraged. For fair comparisons, we provide the initialized results based on the visual features reported as ``Ours(Init)'' and further implement another variant of our method with only visual features available for training, denoted as ``Ours(Vis)''. The performance decrease of ``Ours(Vis)'' proves the contribution and effectiveness of the semantic attributes for the open-set domain adaptation. Moreover, our proposed method reaches promising results on the unseen classes while keeping performance on the seen classes for all tasks. For example, STA achieves the best overall accuracy on task P $\rightarrow$ A, but completely fails on the unseen categories and overfitting to the seen classes. Such an observation emphasizes the superiority of our method in exploring target domain unseen categories.

Table~\ref{table:D2AwA_semantic} show the semantic recovery accuracy on \textit{D2AwA} and \textit{I2AwA}, respectively. Within the expectation, all ZSL methods fail to recognize the data from unseen categories and overfit to the seen classes due to lack of the capacity on tackling the open-set setting. Our proposed method achieves promising results on recognizing both seen and unseen categories, e.g., our method achieves $37.8\%$ accuracy for unseen classes data while keeping $95.2\%$ performance on seen classes for task R$\rightarrow$A. Moreover, our proposed method even outperforms the ABP* and TF-VAE*. They have access to both the seen and unseen categorical attributes in source and target   domain, while our method only employs the seen categories attributes information in the source domain.

\subsection{In-Depth Factor Exploration}
In this subsection, we first visualize the representation from our model, explore the ablation study of the proposed method, showcase several representative samples with the predicted attributes and finally provide more details on the seen and unseen target categories by confusion matrix.

\vspace{1mm}\noindent\textbf{Representation Visualization}. We show the t-SNE embeddings of \textit{I2AwA} from different models in Figure~\ref{fig:tsne}, where red circles denote source data, blue and gray triangles denote target domain seen and unseen classes, respectively. The embedding of our method shows that the same class samples across domains are more compact while discriminative inter classes than the representation produced by source only ResNet-50 \cite{he2016deep} and STA \cite{liu2019separate}. Moreover, our embedding shows the joint visual-semantic representations with more discriminative distribution and separates the unseen categories from seen classes more clear. Such an observation demonstrates the effectiveness of the semantic attributes, which is not only beneficial to the unseen categories, but also promotes the quality of features of the seen classes.

\vspace{1mm}\noindent\textbf{Ablation Study}. 
We dive into our complete method and several variants for open-set domain adaptation and semantic recovery tasks to understand the contribution of each specific design in our framework. As shown in Figure~\ref{fig:terms}, we have the following observations. (1) Compared to w/o R which removes the structure preserving partial alignment term $\mathcal{L}^R$, our method achieves significant performance gains
on the open-set domain task, especially for the seen categories. This demonstrates the effectiveness of aligning the source data to the target domain while preserving the target data structural characteristics. (2) Our method improves the performance D on both tasks compared to w/o, which removes the binary classifier $D(\cdot)$ and only uses classifier $C(\cdot)$ to recognize seen/unseen categories. We conclude that the binary classifier can refine the separation of seen and unseen classes. (3) By removing the attributes propagation mechanism, the performance w/o decreases significantly on the semantic recovery tasks, especially for the unseen categories, proving the contribution of attributes propagation for semantic recovery tasks and uncovering unseen classes. (4) Our method outperforms the variant without constructing visual-semantic fusion w/o VS, which only uses visual features for prediction. For both open-set domain adaptation seen classes and semantic recovery unseen classes, validating the effectiveness of semantic knowledge to the visual features in both preserving performance on seen classes and exploring unseen categories.

\vspace{1mm}\noindent\textbf{Qualitative Demonstration}. To qualitatively illustrate the effectiveness of our method in discovering novel classes and recovering missing semantic information, we further show several representative samples from the target domain unseen categories on \textit{I2AwA} in Figure~\ref{fig:samples}. For each sample, we show some of the correct and wrong predicted attributes with corresponding prediction probabilities. ``P'' and ``R'' indicate the precision and recall score of predicting attributes of each sample. Moreover, some predicted attributes are wrong for the corresponding category, but reasonable for the specific image. From the results, we demonstrate the ability of our model in transferring semantic knowledge from the source domain into the target data, and discovering novel classes through missing semantic information recovery. 

\vspace{1mm}\noindent\textbf{Confusion Matrix}. We visualize the confusion matrix of STA and our method on \textit{I2AwA} in Figure~\ref{fig:confusion}. STA only recognizes those target samples from unseen categories as unknown. On the contrary, our proposed method can discover novel categories in the target domain. Surprisingly, the accuracy of our method for the category ``Giraffe'' achieves $96.5\%$. Moreover, we also notice that not just benefiting uncover unseen categories, our method also enhances the accuracy of the seen classes compared to STA.

\section{Conclusion}

We addressed a novel and practical \emph{Semantic Recovery Open-set Domain Adaptation} problem, which aimed to discover target samples from classes unobserved in the source domain and interpreted based on recovered semantic attributed. To this end, we proposed a novel framework consisting of structural preserving partial alignment, attributes propagation via visual graph, and task-driven classification over joint visual-semantic representations. Finally, two semantic open-set domain adaptation benchmarks were constructed to evaluate our model in terms of open-set recognition and semantic attribute recovery. 

\clearpage
\balance
{\small
\bibliographystyle{ieee_fullname}
\bibliography{egbib}
}

\end{document}

%% file: notation.tex
\begin{table}[t]
\begin{center}\caption{Notations and Descriptions}
\label{table:notation}
\renewcommand{\arraystretch}{1.05}
\footnotesize
\begin{tabular}{cl}

\hline
  Notation & Description\\
  \hline
    $\mathcal{D}_s, \mathcal{D}_t$ & source / target domain \\
    
    $\mathcal{D}_t^s, \mathcal{D}_t^u$ & seen/unseen set\\
    
    $K_s, K_t$ & source / target domain number of classes  \\
    
    $\mathbf{X}_s, \mathbf{X}_t$ & source / target input features \\
    
    $N_s, N_t$ & source / target samples number \\
    
    $N_t^s, N_t^u$ & seen / unseen set samples number \\
    
    $\mathbf{Y}_s, \mathbf{A}_s$  & source domain labels / attributes\\
    
    
    
    $\mathbf{x}_s^i, \mathbf{x}_t^j$ & source / target domain instance \\

    $\mathbf{z}_s^i, \mathbf{z}_t^j$ & source / target domain embedding features \\
    
    
    $\hat{y}_s^i, \hat{y}_t^j$ & predicted source / target label \\
    
   $\hat{a}_s^i, \hat{a}_t^j$ & predicted source / target attributes \\
   
    
    $\mathcal{R}_{\mathbf{x}}, \mathcal{R}_{\mathbf{z}},$ & visual / embedding features prototypes\\
    
    

    
    
    
    $\mathcal{F}_s^i, \mathcal{F}_t^j $ & source / target joint representations\\
   

\hline
\end{tabular}
\end{center}\vspace{-8mm}
\end{table}

%% file: dataset.tex
\begin{table}[t]
\caption{Statistical characteristics on \textit{D2AwA} and \textit{I2AwA} dataset} 
\vspace{1mm}
\label{table:data}
\linespread{1.3} 
\centering
\scriptsize
\setlength{\tabcolsep}{3pt} 
\renewcommand{\arraystretch}{1} 

\begin{tabular}{c|cccccc|cccc}

\hline
Dataset & \multicolumn{6}{c}{\textit{\textit{D2AwA}}} & \multicolumn{2}{|c}{\textit{\textit{I2AwA}}} \\
\hline
 Domain & \multicolumn{2}{c}{\textbf{A}} & \multicolumn{2}{c}{\textbf{P}}& \multicolumn{2}{c}{\textbf{R}}& \multicolumn{1}{|c}{\textbf{I}}& \multicolumn{1}{c}{\textbf{Aw}}\\

\hline
Role & source & target & source & target & source & target & source & target \\
\hline

\#Images  & 9,343 & 16,306& 3,441& 5,760& 5,251& 10,047 & 2,970 & 37,322\\
\#Attributes & 85 & 85 & 85 & 85 & 85 & 85 & 85 & 85 \\
\#Classes & 10 & 17 & 10 & 17 & 10 & 17 & 40 & 50 \\
\hline
\end{tabular}
\vspace{-4mm}
\end{table}

%% file: D2AwA_open.tex
\begin{table*}[t]
\caption{Open-set domain adaptation accuracy ($\%$) on \textit{D2AwA} and \textit{I2AwA}} 
\label{table:D2AwA_open}
\linespread{1.3} 
\centering
\scriptsize
\setlength{\tabcolsep}{4pt} 
\renewcommand{\arraystretch}{1} 

\begin{tabular}{c|cccccccccccccccccc|ccc}

\hline
Dataset & \multicolumn{18}{c}{\textit{\textit{D2AwA}}} & \multicolumn{3}{|c}{\textit{\textit{I2AwA}}} \\
\hline

 Task & \multicolumn{3}{c}{\textbf{A$\rightarrow$P}} & \multicolumn{3}{c}{\textbf{A$\rightarrow$R}} & \multicolumn{3}{c}{\textbf{P$\rightarrow$A}} & \multicolumn{3}{c}{\textbf{P$\rightarrow$R}} & \multicolumn{3}{c}{\textbf{R$\rightarrow$A}} & \multicolumn{3}{c}{\textbf{R$\rightarrow$P}} &
 \multicolumn{3}{|c}{\textbf{I$\rightarrow$Aw}} \\
\hline
Method & OS$^*$ & OS$^\diamond$ & OS & OS$^*$ & OS$^\diamond$ & OS & OS$^*$ & OS$^\diamond$ & OS & OS$^*$ & OS$^\diamond$ & OS & OS$^*$ & OS$^\diamond$ & OS & OS$^*$ & OS$^\diamond$ & OS & OS$^*$ & OS$^\diamond$ & OS\\
\hline

OSBP \cite{saito2018open} & 49.6 & 10.8 & 46.0 & 74.2 & 13.6 & 68.7 &	 76.0 & 9.1 & 69.9 & 63.3 & 6.9 & 58.2 & 90.1 & 13.7 & 83.2 & 55.9 & 10.6 & 51.7 & 67.6 & 7.5 & 66.2\\
STA \cite{liu2019separate} & 60.1 & 33.0 & 57.6 & 85.5 & 10.8 & 78.7 & \textbf{90.2} & 5.7 & \textbf{82.5} & \textbf{82.8} & 7.4 & 76.0 & 88.5 & 7.2 & 81.1 & \textbf{66.9} & 13.5 & 62.0 & 51.5 & 45.5 & 51.4\\
AOD \cite{feng2019attract} & 50.7 & 9.5 & 46.9 & 78.4 & 12.7 & 72.4 & 80.3 & 5.1 & 73.5 & 79.7 & 5.3 & 73.0 & 92.0 & 12.8 & 84.8 & 61.2 & 9.6 & 56.5 & 75.2 & 6.3 & 73.5\\

\hline
Ours(Init) & 53.1 & 45.1 & 52.3 & 78.8 & \textbf{72.3} & 78.2 & 75.3 & 94.8 & 77.1 & 67.3 & 82.0 & 68.6 & 86.2 & 87.7 & 86.4 & 52.0 & 77.8 & 54.4 & 82.2 & 6.3 & 73.5\\

Ours(Vis) & 54.1 & \textbf{76.1} & 56.1 & 75.4 & 70.3 & 75.0 & 69.5 & \textbf{98.5} & 72.1 & 57.4 & 83.1 & 59.7 & 88.3 & \textbf{98.8} & 89.2 & 58.7 & \textbf{91.2} & 61.6& 48.2 & \textbf{70.3} & 48.7\\

Ours & \textbf{62.8} & 47.2 & \textbf{61.4} & \textbf{90.9} &71.4 & \textbf{89.1} & 79.2 & \textbf{98.5} & 81.0 & 78.3 & \textbf{83.7} & \textbf{78.8} & \textbf{94.9} & 90.5 & \textbf{94.5} & 61.2 & 80.4 & \textbf{63.0} & \textbf{83.2} & 70.2 & \textbf{82.8}\\
\hline
\end{tabular}

\end{table*}

%% file: D2AwA_semantic.tex
\begin{table*}[t]
\caption{Semantic Recovery Accuracy ($\%$) on \textit{D2AwA}  and \textit{I2AwA}}
\label{table:D2AwA_semantic}
\linespread{1.3} 
\centering
\scriptsize
\setlength{\tabcolsep}{4pt} 
\renewcommand{\arraystretch}{1} 
\begin{tabular}{c|cccccccccccccccccc|ccc}

\hline
Dataset & \multicolumn{18}{c}{\textit{\textit{D2AwA}}} & \multicolumn{3}{|c}{\textit{\textit{I2AwA}}} \\
\hline

 Task & \multicolumn{3}{c}{\textbf{A$\rightarrow$P}} & \multicolumn{3}{c}{\textbf{A$\rightarrow$R}} & \multicolumn{3}{c}{\textbf{P$\rightarrow$A}} & \multicolumn{3}{c}{\textbf{P$\rightarrow$R}} & \multicolumn{3}{c}{\textbf{R$\rightarrow$A}} & \multicolumn{3}{c}{\textbf{R$\rightarrow$P}} & \multicolumn{3}{|c}{\textbf{I$\rightarrow$Aw}}\\
\hline
Method & S & U & H &S & U & H &S & U & H &S & U & H &S & U & H &S & U & H &S & U & H\\
\hline


\hline
 Source-only	 & 67.6 & 0.0 & 0.0 & 87.6 & 0.0 & 0.0 & 91.3 & 0.0 & 0.0 & \textbf{85.3} & 0.0 & 0.0 & 94.1 & 0.0 & 0.0 & 71.1 & 0.0 & 0.0 & 77.2 & 0.3 & 0.7 \\
ABP \cite{zhu2019learning} & 68.1 & 0.0 & 0.0 & 87.9 & 0.0 & 0.0 & \textbf{91.7} & 0.0 & 0.0 & 83.6 & 0.0 & 0.0 & 94.4 & 0.0 & 0.0 & 70.0 & 0.0 & 0.0 & 79.8 & 0.0 & 0.0\\
TF-VAE \cite{narayan2020latent}&	\textbf{70.4} & 0.0 & 0.0 & 88.4 & 0.0 & 0.0 & 85.1 & 0.0 & 0.0 & 79.6 & 0.0 & 0.0 & \textbf{96.4} & 0.0 & 0.0 & \textbf{72.5} & 0.0 & 0.0 & 62.8 & 0.0 & 0.0\\
\hline
ABP* \cite{zhu2019learning}&64.5 & 6.4 & 11.7 & 86.0 & 5.9 & 11.1 & 84.0 & 24.4 & 37.8 & 81.3 & 12.7 & 21.9 & 93.8 & 16.2 & 27.6 & 67.6 & 7.9 & 14.1 & 78.0 & 13.4 & 22.9 \\
TF-VAE* \cite{narayan2020latent} &59.7 & 12.8 & 21.0 & 77.9 & 16.4 & 27.1 & 35.1 & 35.6 & 35.3 & 34.8 & \textbf{32.7} & \textbf{33.7} & 68.5 & 36.1 & 47.3 & 50.7 & \textbf{21.0} & 29.7 & 37.7 & 20.0 & 26.2\\

\hline
Ours & 62.5 & \textbf{27.0} & \textbf{37.7} & \textbf{90.7} & \textbf{30.0} & \textbf{45.1} & 79.2 & \textbf{36.7} & \textbf{50.2} & 78.0 & 15.7 & 26.1 & 95.2 & \textbf{37.8} & \textbf{54.1} & 59.0 & 20.8 & \textbf{30.8} & \textbf{83.1} & \textbf{22.0} & \textbf{34.8}\\
\hline
\end{tabular}
\vspace{-2mm}
\end{table*}